\documentclass[sigconf]{acmart}
\usepackage{makecell}
\usepackage{multirow}
\usepackage{graphicx}
\usepackage{subfigure}
\usepackage[normalem]{ulem}
\usepackage{url}
\usepackage{enumerate}
\usepackage{booktabs}
\usepackage{multirow}
\usepackage{xspace}

\usepackage{amsmath}
\usepackage{color}
\usepackage{threeparttable}
\usepackage{ulem}
\usepackage{amssymb}

\usepackage{float}
\usepackage[ruled,linesnumbered]{algorithm2e}
\usepackage{algorithm2e, setspace, algpseudocode}
\usepackage{colortbl}

\newcommand{\boldres}[1]{{\textbf{{#1}}}}

\AtBeginDocument{%
  \providecommand\BibTeX{{%
    \normalfont B\kern-0.5em{\scshape i\kern-0.25em b}\kern-0.8em\TeX}}}

\copyrightyear{2024}
\acmYear{2024}
\setcopyright{rightsretained}
\acmConference[CIKM '24]{Proceedings of the 33rd ACM International
Conference on Information and Knowledge Management}{October 21--25,
2024}{Boise, ID, USA}
\acmBooktitle{Proceedings of the 33rd ACM International Conference on
Information and Knowledge Management (CIKM '24), October 21--25, 2024,
Boise, ID, USA}
\acmDOI{10.1145/3627673.3679884}
\acmISBN{979-8-4007-0436-9/24/10}

\usepackage{etoolbox}
\makeatletter
\patchcmd{\maketitle}{\@copyrightpermission}{
$^\dagger$ Also with The Hong Kong Polytechnic University. $^\ddagger$ Corresponding author.\\
   \begin{minipage}{0.3\columnwidth}
     \href{https://creativecommons.org/licenses/by/4.0/}{\includegraphics[width=0.90\textwidth]{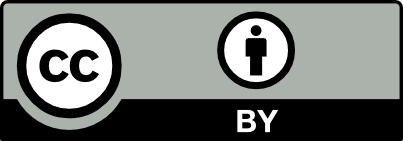}}
   \end{minipage}\hfill
   \begin{minipage}{0.7\columnwidth}
     \href{https://creativecommons.org/licenses/by/4.0/}{
     This work is licensed under a Creative Commons Attribution International 4.0 License.}
   \end{minipage}
  
   \vspace{5pt}
}{}{}

\makeatother

\begin{document}

\title{Channel-Aware Low-Rank Adaptation in Time Series Forecasting}
%%
%% By default, the full list of authors will be used in the page
%% headers. Often, this list is too long, and will overlap
%% other information printed in the page headers. This command allows
%% the author to define a more concise list
%% of authors' names for this purpose.

\author{Tong Nie$^\dagger$}
\affiliation{%
  \institution{Tongji University}
  \city{Shanghai}
  \country{China}}
% \affiliation{%
%   \institution{The Hong Kong Polytechnic University}
%   \city{Hong Kong SAR}
%   \country{China}}
\email{nietong@tongji.edu.cn}

\author{Yuewen Mei}
\affiliation{%
  \institution{Tongji University}
  \city{Shanghai}
  \country{China}}
\email{meiyuewen@tongji.edu.cn}

\author{Guoyang Qin}
\affiliation{%
  \institution{Tongji University}
  \city{Shanghai}
  \country{China}}
\email{2015qgy@tongji.edu.cn}

\author{Jian Sun}
\affiliation{%
  \institution{Tongji University}
  \city{Shanghai}
  \country{China}}
\email{sunjian@tongji.edu.cn}

\author{Wei Ma$^\ddagger$}
\affiliation{%
  \institution{The Hong Kong Polytechnic University}
  \city{Hong Kong SAR}
  \country{China}
}
\email{wei.w.ma@polyu.edu.hk}

\renewcommand{\authors}{Tong Nie, Yuewen Mei, Guoyang Qin, Jian Sun, and Wei Ma}
\renewcommand{\shortauthors}{Tong Nie, Yuewen Mei, Guoyang Qin, Jian Sun, \& Wei Ma}

%%
%% The abstract is a short summary of the work to be presented in the
%% article.
\begin{abstract}
The balance between model capacity and generalization has been a key focus of recent discussions in long-term time series forecasting. Two representative channel strategies are closely associated with model expressivity and robustness, including channel independence (CI) and channel dependence (CD). The former adopts individual channel treatment and has been shown to be more robust to distribution shifts, but lacks sufficient capacity to model meaningful channel interactions. The latter is more expressive for representing complex cross-channel dependencies, but is prone to overfitting. To balance the two strategies, we present a channel-aware low-rank adaptation method to condition CD models on identity-aware individual components. As a plug-in solution, it is adaptable for a wide range of backbone architectures. Extensive experiments show that it can consistently and significantly improve the performance of both CI and CD models with demonstrated efficiency and flexibility. The code is available at \underline{\url{https://github.com/tongnie/C-LoRA}}.
\end{abstract}

%%
%% The code below is generated by the tool at http://dl.acm.org/ccs.cfm.
%% Please copy and paste the code instead of the example below.
%%
\begin{CCSXML}
<ccs2012>
   <concept>
       <concept_id>10002951.10003227.10003351</concept_id>
       <concept_desc>Information systems~Data mining</concept_desc>
       <concept_significance>500</concept_significance>
       </concept>
 </ccs2012>
\end{CCSXML}

\ccsdesc[500]{Information systems~Data mining}

%%
%% Keywords. The author(s) should pick words that accurately describe
%% the work being presented. Separate the keywords with commas.
\keywords{Long-Term Time Series Forecasting, Low-Rank Adaptation, Channel Independence, Channel Mixing}

%%
%% This command processes the author and affiliation and title
%% information and builds the first part of the formatted document.
\maketitle

\section{Introduction}

% \textbf{The progress of long-term series forecasting models, introduce the two branches of models: Transformer-based and linear models.}

As a fundamental scientific problem in diverse fields such as weather, economics, energy, and transportation, multivariate time series forecasting has attracted great interest in recent years. Especially, advanced neural forecasting architectures are designed to address the challenging long-term series forecasting (LTSF) problem, and two branches of methods, including Transformer- and MLP-based models, have achieved remarkable progress \cite{wu2021autoformer,zhang2022crossformer,zeng2023transformers,chen2023tsmixer,liu2023itransformer,wang2023timemixer,nie2023contextualizing,yi2024frequency}.

\vspace{-0.3cm}
\begin{figure}[!htbp]
  \centering
  \captionsetup{skip=1pt}
  \includegraphics[width=1\columnwidth]{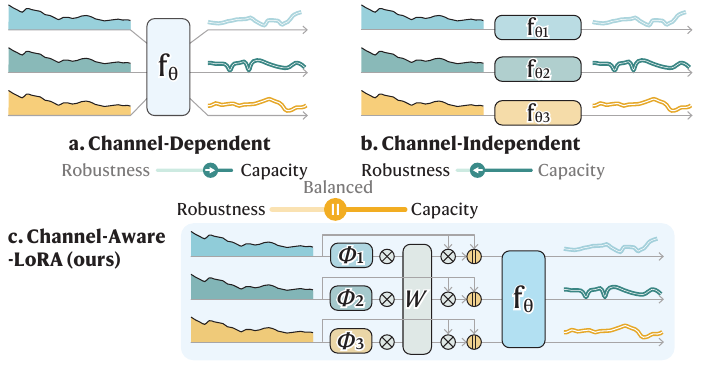}
  \caption{The proposed channel-aware low-rank adaptation.}
  \label{fig:intro}
\end{figure}
\vspace{-0.3cm}

% \vspace{-0.4cm}
% \textbf{The challenge faced in LTSF: balance between generalization and capacity.}
By exploring the two types of architectures, recent studies have identified the core challenges faced in LTSF: \textit{the balance between generalization and model capacity} \cite{han2024capacity}. On the one hand, real-world time series can be non-stationary, manifesting as a distribution shift in both training and test data. For example, the hourly energy consumption can vary in different workdays and seasons.
Models need to be robust to confront the biased distribution.
On the other hand, LTSF models are required to have enough capacity to represent the interweaved relationships between channels and possibly heterogeneous channel patterns. This frequently occurs in real-world time series, such as the weather and traffic system.

% 
% \textbf{Introduce the channel-independent models (especially the channel-individual models) and the channel-mixing models. The shortcomings of both classes.}
To address these challenges, very recent studies have shifted the focus from architecture designs to discussions of effective channel strategies \cite{nie2022time,zeng2023transformers,liu2023itransformer,chen2024similarity}, as channel management methods can impact both robustness and expressivity.
Generally, there are two primary channel methods: Channel-Independent (CI) and Channel-Dependent (CD) strategies. CD models treat series in all channels as a whole and employ a global predictor to model channel relations implicitly or explicitly.
In contrast, CI strategy views multivariate channels as multiple univariate series to consider fine-grained individual patterns and overlooks potential channel interactions. Empirical results have shown that CI models are more robust in addressing distribution shift but have too expensive hypothesis spaces; while CD models have larger capacities but are prone to overfit and less flexible in modeling channel-specific phenomena.
\textbf{Limitation of existing approaches.}
The above dilemma calls for a new channel strategy that can balance CI and CD methods. Several pioneering approaches are leading the way in this field, such as group-aware embedding \cite{xiao2023gaformer} and inverted embedding \cite{liu2023itransformer} for Transformers, leading indicator estimation \cite{zhao2024rethinking}, and channel clustering \cite{chen2024similarity}.
However, they are either limited to specific types of backbone model, or hard to improve the performance by a large margin without significantly increasing the computational burden.

\textbf{Our solution.}
Inspired by recent progress in low-rank adaptation \cite{hu2021lora}, we propose a channel-aware low-rank adaptation (C-LoRA) method to achieve a trade-off between the two strategies and provide an alternative in a parameter-efficient way. Specifically, we parameterize each channel a low-rank factorized adapter to consider individual treatment. Then the specialized channel adaptation is conditioned on the series information to form an identity-aware embedding. The cross-channel relational dependencies are simultaneously exploited by combining a globally shared CD model.

\textbf{Contribution.}
C-LoRA is a plug-in solution that is seamlessly adaptable to a wide range of SOTA time series models. It contains almost no changes to the existing architecture. Extensive experiments demonstrate that it can consistently improve the performance of both CD and CI backbones by a large margin. Moreover, it has great efficiency, flexibility to transfer across datasets, and can enhance channel identity for accurate channel interaction modeling.

\section{Preliminaries}
\textbf{Multivariate Time Series Forecasting.}
Given the historical set $\mathbf{X}=\{\mathbf{x}_1,\dots\mathbf{x}_T\}\in\mathbb{R}^{T\times C}$, where $T$ is the look-back window and $C$ is the number of channels (variates), our objective is to learn a predictive function $f_{\theta}$ to estimate the future set $\mathbf{Y}=\{\mathbf{x}_{T+1},\dots,\mathbf{x}_{T+H}\}\in\mathbb{R}^{H\times C}$ with a horizon $H$. For brevity, we denote $\mathbf{X}_{:,c}\in\mathbb{R}^{T}$ as the $c$-th channel and $\mathbf{X}_{t,:}\in\mathbb{R}^{C}$ as the observations at time $t$.

\noindent\textbf{Channel-Dependent (CD).} 
CD strategy is widely adopted in multivariate time series forecasting. CD models treat series in all channels as a whole and employ a global predictor $f_\theta:\mathbb{R}^{T\times C}\mapsto\mathbb{R}^{H\times C}$ to simultaneously forecast all components. In particular, a special case is called \textit{channel mixing (CM)}, which models the explicit channel information interaction either by MLPs or self-attention.
Given the training set $\{(\mathbf{X}^{(i)},\mathbf{Y}^{(i)})\}_{i=1}^N$, the global CD loss is given by:
\begin{equation}\label{eq:cd_loss}
    \theta^*=\arg \min_{\theta}\frac{1}{N}\sum_{i=1}^N\Vert \mathbf{Y}^{(i)}-f_{\theta}(\mathbf{X}^{(i)})\Vert_F^2,
\end{equation}
where $N$ is the number of training samples.

\noindent\textbf{Channel-Independent (CI).} 
In contrast, CI strategy views multivariate channels as multiple univariate time series and ignores channel interactions. Particularly, the \textit{channel-individual (CInd)} model is a special case that assigns a unique local predictor for each channel: $f_{\theta_i}:\mathbb{R}^{T}\mapsto\mathbb{R}^{H}, \forall i=1,\dots,C$. The CI loss is given by:
\begin{equation}
    \theta_1^*,\dots,\theta_C^*=\arg \min_{\theta_1,\dots,\theta_C}\frac{1}{NC}\sum_{i=1}^N\sum_{c=1}^C\Vert \mathbf{Y}^{(i)}_{:,c}-f_{\theta_c}(\mathbf{X}^{(i)}_{:,c})\Vert_F^2.
\end{equation}
% where the loss is averaged over all channels and each channel is optimized independently.
\section{Methodology}
As stated above, CI models are more robust in addressing distribution shift but have too expensive hypothesis spaces; while CD models have larger model capacities but are prone to overfit \cite{han2024capacity}. This section proposes the channel-aware low-rank adaptation model to balance channel-specific treatment and channel-wise dependence.

\vspace{-0.2cm}
\subsection{General Forecasting Backbone}
As an efficient plugin, C-LoRA can be easily incorporated into a wide range of forecasting architectures. We first introduce a general forecasting template for both the CI and CD models as follows.
\begin{equation}
\begin{aligned}
    &\widebar{\mathbf{X}}=\textsc{Normalization}(\mathbf{X}), \\
    &\mathbf{z}^{(0)}_{c}=\textsc{TokenEmbedding}(\widebar{\mathbf{X}}_{:,c}), \forall c=1,\dots, C,\\
    \text{(Optional):}~& \mathbf{Z}^{(\ell+1)}=\textsc{ChannelMixing}(\mathbf{Z}^{(\ell)}), \forall \ell=0,\dots, L,\\
    &\widehat{\mathbf{Y}}=\textsc{Projection}(\mathbf{Z}^{(L+1)}), \\
\end{aligned}
\end{equation}
where $\textsc{Normalization}$ such as RevIN \cite{kim2021reversible} is adopted to address the nonstationarity of time series, $\textsc{TokenEmbedding}: \mathbb{R}^{T}\mapsto\mathbb{R}^{D}$ and $\textsc{Projection}: \mathbb{R}^{D}\mapsto\mathbb{R}^{H}$ are usually implemented by MLPs to process temporal features, and $\textsc{ChannelMixing}: \mathbb{R}^{C\times D}\mapsto\mathbb{R}^{C\times D}$ is optional for CD models by Transformer blocks or MLPs.

Many state-of-the-art forecasting frameworks follow this template, such as TSMixer \cite{chen2023tsmixer}, RMLP \cite{li2023revisiting}, iTransformer \cite{liu2023itransformer}, and FreTS \cite{yi2024frequency}. For other Transformer-based architectures that employ token-wise attention, such as Autoformer \cite{wu2021autoformer} and Informer \cite{zhou2021informer}, we can adapt them with a simple inverting strategy \cite{liu2023itransformer} and adopt C-LoRA.

\vspace{-0.2cm}
\subsection{Channel-Aware Low-Rank Adaptation}
We elaborate C-LoRA by first revisiting the two strategies.
To account for channel-specific effects, CI (or CInd) methods can adopt individual models for each channel, which instantiate the \textsc{TokenEmbedding} with a series of mappings, e.g., different MLPs:
\begin{equation}
    \mathbf{z}^{(0)}_{c}=\texttt{MLP}_c(\widebar{\mathbf{X}}_{:,c}; \theta_c), \forall c=1,\dots, C,
\end{equation}
then the hypothesis class of all individuals is $\mathcal{H}_{\text{CI}}=\{\texttt{MLP}_c(\cdot; \theta_c)|\theta_c\in\Theta,c=1,\dots,C\}$ where $\Theta$ is the parameter space. However, such a hypothesis class is computationally expensive, and pure CI models fail to exploit multivariate correlational structures \cite{cini2024taming}. 

In contrast, the CD strategy is more expressive by modeling channel interactions either explicitly with \textsc{ChannelMixing} or implicitly by optimizing the global loss in Eq. \eqref{eq:cd_loss}.
However, they can have difficulty capturing individual channel patterns with a shared encoder $\texttt{MLP}(\widebar{\mathbf{X}}; \theta)$, and the CM operation can generate mixed channel identity information, causing an indistinguishment issue \cite{nie2023contextualizing}.

Combining the merits of the two strategies, we propose a novel strategy that considers channel-wise adaptation in a CD model. Specifically, to model individual channels in a parameter-efficient way, a low-rank \textit{adapter} is specialized for each channel $\phi^{(c)}\in\mathbb{R}^{r\times D}$, where $r\ll D$ is the intrinsic rank. Then we can condition on another low-rank matrix to project it to a larger dimension:
\begin{equation}\label{eq:clora_form}
    \widetilde{\phi}^{(c)}=\texttt{ReLU}(\phi^{(c),\mathsf{T}}\mathbf{W})\in\mathbb{R}^{D\times d},
\end{equation}
where $\mathbf{W}\in\mathbb{R}^{r\times d}$ and $d$ is the adaptation dimension. $\widetilde{\phi}^{(c)}$ characterizes channel-specific parameters, and it needs to be aware of the series information to consider the channel identity. Then we have:
\begin{equation}
    \mathbf{z}^{(0)}_{c,\phi} = \mathbf{z}^{(0),\mathsf{T}}_{c}\widetilde{\phi}^{(c)}\in\mathbb{R}^{d},
\end{equation}
where $\mathbf{z}^{(0)}_{c}=\texttt{MLP}(\widebar{\mathbf{X}}_{:,c}; \theta)$ is obtained by a CD model shared by all channels.
By aggregating all channel adaptations $\mathbf{Z}^{(0)}_{\phi}=\{\mathbf{z}^{(0)}_{c,\phi}\}_{c=1}^C\in\mathbb{R}^{C\times d}$, 
we incorporate it into the global CD models to exploit multivariate correlations. Then, the final C-LoRA is given by:
\begin{equation}
    \mathbf{Z}^{(0)} = \left[\texttt{MLP}(\widebar{\mathbf{X}}; \theta)\Vert\mathbf{Z}^{(0)}_{\phi}\right]\in\mathbb{R}^{C\times (D+d)},
\end{equation}
where $[\cdot]$ is the concatenating operation, and $\mathbf{Z}^{(0)}$ is passed to the next modules. Note that C-LoRA achieves a balance between CD and CI models and efficiently integrates global-local components. It can adapt to individual channels with the specialized channel adaptation $\mathbf{z}^{(0)}_{c,\phi}$, and preserve multivariate interactions by the shared $\texttt{MLP}(\widebar{\mathbf{X}}_{:,c}; \theta)$.
The reduced hypothesis class is $\mathcal{H}_{\text{C-LoRA}}=\{\texttt{MLP}(\cdot;\theta,\phi^{(c)})|\theta\in\Theta,\phi^{(c)}\in\mathbb{R}^{r\times D}\}$, which is more efficient.

% \subsection{Channel Alignment Regularization}

\begin{table*}[!htbp]
  \caption{Results of the LTSF benchmarks. We report the forecast error of different models under different prediction lengths. The input sequence length is set to 96 for all methods. \emph{IMP} shows the average percentage of MSE/MAE improvement of C-LoRA.}
  \label{tab:main}
  \vskip -0.0in
  \vspace{0pt}
  \renewcommand{\arraystretch}{1} 
  \centering
  \resizebox{1\textwidth}{!}{
  \begin{threeparttable}
  \begin{small}
  \renewcommand{\multirowsetup}{\centering}
  \setlength{\tabcolsep}{1pt}
  \begin{tabular}{c|c|cc>{\columncolor{gray!10}}c>{\columncolor{gray!10}}c|cc>{\columncolor{gray!10}}c>{\columncolor{gray!10}}c|cc>{\columncolor{gray!10}}c>{\columncolor{gray!10}}c|cc>{\columncolor{gray!10}}c>{\columncolor{gray!10}}c|cc>{\columncolor{gray!10}}c>{\columncolor{gray!10}}c|cc>{\columncolor{gray!10}}c>{\columncolor{gray!10}}c|cc>{\columncolor{gray!10}}c>{\columncolor{gray!10}}c|>{\columncolor{gray!20}}c}
    \toprule
    \multicolumn{2}{c}{\multirow{2}{*}{Models}}  &
    \multicolumn{2}{|c}{\rotatebox{0}{{iTransformer}}} &
    \multicolumn{2}{c|}{\rotatebox{0}{{w/ C-LoRA}}} &
    \multicolumn{2}{c}{\rotatebox{0}{{TSMixer}}} &
    \multicolumn{2}{c|}{\rotatebox{0}{{w/ C-LoRA}}} &
    \multicolumn{2}{c}{\rotatebox{0}{{RMLP}}}&
    \multicolumn{2}{c}{\rotatebox{0}{{w/ C-LoRA}}} &
    \multicolumn{2}{|c}{\rotatebox{0}{{FreTS}}} &
    \multicolumn{2}{c|}{\rotatebox{0}{{w/ C-LoRA}}} &
    \multicolumn{2}{c}{\rotatebox{0}{{FEDformer}}} &
    \multicolumn{2}{c|}{\rotatebox{0}{{w/ C-LoRA}}} &
    \multicolumn{2}{c}{\rotatebox{0}{{Autoformer}}} &
    \multicolumn{2}{c}{\rotatebox{0}{{w/ C-LoRA}}} &
    \multicolumn{2}{|c}{\rotatebox{0}{{Informer}}} &
    \multicolumn{2}{c|}{\rotatebox{0}{{w/ C-LoRA}}} & \multicolumn{1}{c}{\rotatebox{0}{IMP}}\\
    \cmidrule(lr){3-4} \cmidrule(lr){4-6}\cmidrule(lr){7-8} \cmidrule(lr){8-10}\cmidrule(lr){11-12}\cmidrule(lr){12-14} \cmidrule(lr){15-16} \cmidrule(lr){16-18} \cmidrule(lr){19-20}  \cmidrule(lr){20-22} \cmidrule(lr){23-24} \cmidrule(lr){24-26}  \cmidrule(lr){26-28}  \cmidrule(lr){28-31} 
    \multicolumn{2}{c}{Metric}   & \multicolumn{1}{|c}{{MSE}} & {MAE}  & {MSE} & {MAE}  & {MSE} & {MAE}  & {MSE} & {MAE}  & {MSE} & {MAE}  & {MSE} & {MAE} & {MSE} & {MAE} & {MSE} & {MAE} & {MSE} & {MAE} & {MSE} & {MAE} & {MSE} & {MAE} & {MSE} & {MAE} & {MSE} & {MAE}& {MSE} & {MAE} & $\%$\\
    \toprule
    \multirow{5}{*}{\rotatebox{90}{{ETTm1}}} 
    &  {96}  &{0.345} &{0.376} &{\boldres{0.331}} &{\boldres{0.367}} &{0.332} &{0.370} &{\boldres{0.317}} &{\boldres{0.356}} &{{0.337}} &{{0.374}} &{{\boldres{0.321}}} &{{\boldres{0.360}}}&{0.340} &{0.376}&{\boldres{0.330}} &{\boldres{0.369}}&{0.379} &{0.419} &{\boldres{0.375}} &{\boldres{0.422}} &{\boldres{0.505}} &{\boldres{0.475}} &{0.512} &{0.480} &{0.672} &{0.571} &{\boldres{0.577}} &{\boldres{0.542}} & {3.23}
    \\
    & {192}  &{0.384} &{0.394} &{\boldres{0.373}} &{\boldres{0.390}} &{0.372} &{0.390} &{\boldres{0.358}} &{\boldres{0.377}} &{{0.379}} &{{0.391}} &{{\boldres{0.363}}} &{{\boldres{0.380}}} &{0.383} &{0.398}&{\boldres{0.375}} &{\boldres{0.398}}&{0.426} &{0.441} &{\boldres{0.411}} &{\boldres{0.435}} &{\boldres{0.553}} &{\boldres{0.496}} &{0.558} &{0.503} &{0.795} &{0.669}&{\boldres{0.720}} &{\boldres{0.634}} & {2.67}
    \\
    & {336} &{0.418} &{0.415} &{\boldres{0.409}} &{\boldres{0.414}}  &{0.405} &{0.411} &{\boldres{0.389}} &{\boldres{0.400}} &{{0.412}} &{{0.412}} &{{\boldres{0.395}}} &{{\boldres{0.403}}} &{0.420} &{0.425}&{\boldres{0.406}} &{\boldres{0.418}}&{0.445} &{0.459} &{\boldres{0.423}} &{\boldres{0.447}} &{0.621} &{0.537} &{\boldres{0.618}} &{\boldres{0.523}} &{1.212} &{0.871}&{\boldres{0.982}} &{\boldres{0.756}} &{4.51}
    \\
    & {720}  &{0.481} &{0.451} &{\boldres{0.479}} &{\boldres{0.449}} &{0.469} &{0.447} &{\boldres{0.455}} &{\boldres{0.436}} &{{0.478}} &{{0.447}} &{{\boldres{0.462}}} &{{\boldres{0.440}}} &{0.499} &{0.477}&{\boldres{0.479}} &{\boldres{0.462}}&{0.543} &{0.490} &{\boldres{0.509}} &{\boldres{0.488}} &{0.671} &{0.561} &{\boldres{0.592}} &{\boldres{0.520}} &{1.166} &{0.823}&{\boldres{1.121}} &{\boldres{0.794}} &{3.68}
    \\
    \cmidrule(lr){2-31}
    & {Avg}  &{0.407} &{0.409} &{\boldres{0.398}} &{\boldres{0.405}} &{0.395} &{0.405} &{\boldres{0.380}} &{\boldres{0.392}} &{0.402} &{0.406} &{\boldres{0.385}} &{\boldres{0.396}} &{0.411} &{0.419} &{\boldres{0.398}} &{\boldres{0.412}} &{0.448} &{0.452} &{\boldres{0.430}} &\boldres{{0.448}} &{0.588} &{0.517} &{\boldres{0.570}} &{\boldres{0.507}} &{0.961} &{0.734} &{\boldres{0.850}} &{\boldres{0.682}}&{3.59}
    \\
    \midrule
    \multirow{5}{*}{\rotatebox{90}{{ETTh1}}} 
    &  {96}  &{\boldres{0.390}} &{\boldres{0.406}} &{\boldres{0.390}} &{0.407} &{0.398} &{0.412} &{\boldres{0.396}} &{\boldres{0.409}} &{{0.405}} &{{0.413}} & {\boldres{0.381}} &{{\boldres{0.394}}} &{\boldres{0.398}} &{0.410}&{0.402} &{\boldres{0.409}}&{{\boldres{0.376}}} &{0.419} &{\boldres{0.376}} &{\boldres{0.417}} &{\boldres{0.449}} &{\boldres{0.459}} &{0.453} &{0.469} &{0.912} &{0.717} &{\boldres{0.874}} &{\boldres{0.715}} &{0.90}
    \\
    & {192}  &{0.442} &{\boldres{0.436}} &{\boldres{0.440}} &{\boldres{0.436}} &{\boldres{0.451}} &{0.442} &{\boldres{0.451}} &{\boldres{0.440}} &{{0.460}} &{{0.444}} &{\boldres{0.442}} &{{\boldres{0.427}}}  &{\boldres{0.453}} &{0.446}&{0.460} &{\boldres{0.444}}&{{0.420}} &{0.448} &{\boldres{0.419}} &{\boldres{0.443}} &{0.500} &{0.482} &{\boldres{0.490}} &{\boldres{0.477}} &{\boldres{1.037}} &{0.779} &{{1.044}} &{\boldres{0.775}} &{0.84}
    \\
    & {336}  &{0.485} &{0.459} &{\boldres{0.482}} &{\boldres{0.457}}  &{\boldres{0.489}} &{0.461} &{{0.493}} &{\boldres{0.460}}  &{0.505} &{0.466} &{{\boldres{0.493}}} & {{\boldres{0.455}}} &{\boldres{0.508}} &{0.480}&{0.509} &{\boldres{0.473}}&{{0.459}} &{{\boldres{0.465}}} &{\boldres{0.458}} &{0.466} &{0.521} &{0.496} &{\boldres{0.511}} &{\boldres{0.487}} &{\boldres{1.116}} &{\boldres{0.816}}  &{1.146} &{{0.843}} &{0.30}
    \\
    & {720} &{0.493} &{0.485} &{\boldres{0.486}} &{\boldres{0.477}} &{0.507} &{0.486} &{\boldres{0.498}} &{\boldres{0.480}} &{0.514} &{{0.490}} &{\boldres{0.493}} &{\boldres{0.475}}&{0.560} &{0.537}&{\boldres{0.547}} &{\boldres{0.515}}  &{{0.506}} &{{0.507}} &{\boldres{0.472}} &{\boldres{0.486}} &{{0.514}} &{0.512} &{\boldres{0.503}} &{\boldres{0.508}} &{1.230} &{0.887} &{\boldres{1.170}} &{\boldres{0.869}} &{2.88}
    \\
    \cmidrule(lr){2-31}
    & {Avg}  &{0.453} &{0.447} &{\boldres{0.450}} &{\boldres{0.444}} &{0.461} &{0.450} &{\boldres{0.460}} &{\boldres{0.447}} &{0.471} &{0.453} &{\boldres{0.452}} &{\boldres{0.438}} &{0.480} &{0.468} &{0.480} &{\boldres{0.460}} &{0.440} &{0.460} &{\boldres{0.431}} &{\boldres{0.453}} &{0.496} &{0.487} &{\boldres{0.489}} &{\boldres{0.485}} &{1.074} &{0.800} &{\boldres{1.059}} &{0.801} &{1.28}
    \\
    \midrule
    \multirow{5}{*}{\rotatebox{90}{{Electricity}}} 
    &  {96}  & {0.148} & {0.240} & {\boldres{0.139}} & {\boldres{0.234}} & {0.177} & {0.278} & {\boldres{0.155}} & {\boldres{0.256}}  &{{0.201}} &{{0.287}} &{\boldres{0.168}} &{\boldres{0.258}}&{0.320} &{0.403}&{\boldres{0.165}} &{\boldres{0.262}} &{0.195} &{0.309} &{{\boldres{0.193}}} &{{\boldres{0.307}}} &{0.203} &{0.318}  &{\boldres{0.193}} &{\boldres{0.307}} &{\boldres{0.329}} &{\boldres{0.407}}  &{\boldres{0.329}} &{0.412}&{10.55}
    \\
    & {192}  & {0.162} & {0.253} & {\boldres{0.160}} & {\boldres{0.254}} & {0.193} & {0.293} & {\boldres{0.172}} & {\boldres{0.270}}  &{{0.209}} &{0.297} &{\boldres{0.179}} &{{\boldres{0.268}}}&{0.325} &{0.406}&{\boldres{0.176}} &{\boldres{0.270}} &{\boldres{0.202}} &{\boldres{0.315}} &{{0.203}} &{{0.316}} &{0.225} &{0.334} &{\boldres{0.222}} &{\boldres{0.330}} &{\boldres{0.338}} &{\boldres{0.419}} &{{0.347}} &{0.430} &{8.53}
    \\
    & {336}  & {0.178} & {0.269} & {\boldres{0.171}} & {\boldres{0.266}} & {0.215} & {0.315} & {\boldres{0.191}} & {\boldres{0.289}}  &{{0.228}} &{{0.316}} &{\boldres{0.196}} &{{\boldres{0.285}}}&{0.370} &{0.435}&{\boldres{0.195}} &{\boldres{0.289}} &{0.234} &{0.347} &{{\boldres{0.230}}} &{\boldres{0.343}} &{0.282} &{0.377}  &{\boldres{0.266}} &{\boldres{0.370}} &{0.364} &{0.439} &{\boldres{0.352}} &{\boldres{0.434}} &{10.29}
    \\
    & {720}  & {0.225} & {0.317} & {\boldres{0.195}} & {\boldres{0.289}} & {0.260} & {0.352} & {\boldres{0.230}} & {\boldres{0.321}}  &{{0.273}} &{{0.350}} &{\boldres{0.238}} &{\boldres{0.319}} &{0.416} &{0.474}&{\boldres{0.235}} &{\boldres{0.325}}&{0.261} &{\boldres{0.365}} &{{0.262}} &{{\boldres{0.365}}} &{0.314} &{\boldres{0.383}} &{\boldres{0.299}} &{0.389} &{0.397} &{0.460} &{\boldres{0.395}} &{\boldres{0.456}} &{10.24}
    \\
    \cmidrule(lr){2-31}
    & {Avg}  & {0.178} & {0.270} & {\boldres{0.166}} & {\boldres{0.261}} & {0.211} & {0.310} & {\boldres{0.187}} & {\boldres{0.284}} & {0.228} & {0.313} & {\boldres{0.195}} & {\boldres{0.283}} & {0.358} & {0.430} & {\boldres{0.193}} & {\boldres{0.287}} & {0.223} & {0.334} & {\boldres{0.222}} & {\boldres{0.333}} & {0.256} & {0.353} & {\boldres{0.245}} & {\boldres{0.349}} & {0.357} & {0.431} & {\boldres{0.356}} & {0.433} & {9.95} 
    \\
    \midrule
    \multirow{5}{*}{\rotatebox{90}{{Weather}}} 
    &  {96}  & {0.174} & {0.214} & {\boldres{0.164}} & {\boldres{0.209}} & {0.181} & {0.228} & {\boldres{0.158}} & {\boldres{0.206}}  &{{0.196}} &{{0.235}} & {\boldres{0.163}} &{\boldres{0.208}}&{0.186} &{0.241}&{\boldres{0.165}} &{\boldres{0.228}} & {0.220} &{0.300} & {{\boldres{0.218}}} &{{\boldres{0.299}}} & {0.266} &{0.336}&{\boldres{0.234}} &{\boldres{0.312}} & {0.300} &{0.384} &{\boldres{0.265}} &{\boldres{0.348}} &{8.35}
    \\
    & {192}  & {0.221} & {0.254} & {\boldres{0.209}} & {\boldres{0.251}} & {0.227} & {0.263} & {\boldres{0.207}} & {\boldres{0.249}}  &{{0.240}} &{{0.271}}  & {\boldres{0.209}} &{\boldres{0.249}} &{0.222} &{\boldres{0.273}}&{\boldres{0.210}} &{{0.274}}& {\boldres{0.278}} &{\boldres{0.344}} & {0.282} &{0.350} & {0.307} &{0.367} &{\boldres{0.282}} &{\boldres{0.344}} & {0.598} &{0.544} &{\boldres{0.381}} &{\boldres{0.427}} &{8.27}
    \\
    & {336}  & {0.278} & {0.296} & {\boldres{0.268}} & {\boldres{0.294}} & {0.280} & {0.300} & {\boldres{0.266}} & {\boldres{0.292}}  &{{0.291}} &{{0.307}} & {\boldres{0.264}} &{\boldres{0.289}}&{0.272} &{0.316}&{\boldres{0.264}} &{\boldres{0.317}} & {\boldres{0.339}} &{\boldres{0.382}} & {0.349} &{0.390} & {0.359} &{\boldres{0.395}} &{\boldres{0.357}} &{\boldres{0.395}} &{0.578} &{0.523} &{\boldres{0.515}} &{\boldres{0.511}} &{2.74}
    \\
    & {720}  & {0.358} & {0.349} & {\boldres{0.349}} & {\boldres{0.346}} & {0.353} & {0.347} & {\boldres{0.348}} & {\boldres{0.345}}  &{0.363} &{{0.353}} & {{\boldres{0.342}}} &{{\boldres{0.340}}} &{0.350} &{0.381}&{\boldres{0.343}} &{\boldres{0.370}}& {0.409} &{0.438} & {0.411} &{\boldres{0.420}} & {0.419} &{0.428} &{\boldres{0.415}} &{\boldres{0.424}} & {1.059} &{0.741} &{\boldres{0.792}} &{\boldres{0.651}} &{4.47}
    \\
    \cmidrule(lr){2-31}
    & {Avg}  & {0.258} & {0.278} & {\boldres{0.248}} & {\boldres{0.275}} & {0.260} & {0.285} & {\boldres{0.245}}& {\boldres{0.273}} & {0.273} & {0.292} & {\boldres{0.245}} & {\boldres{0.272}} & {0.258} & {0.303} & {\boldres{0.246}} & {\boldres{0.297}} & {0.312} & {0.366} & {0.315} & {\boldres{0.365}} & {0.338} & {0.382} & {\boldres{0.322}} & {\boldres{0.369}} & {0.634} & {0.548} & {\boldres{0.488}} & {\boldres{0.484}} & {5.76}
    \\
    \midrule
    \multirow{5}{*}{\rotatebox{90}{{Solar}}} 
    &  {96}  &{0.203} &{0.237} & {\boldres{0.175}} & {\boldres{0.219}} &{0.222} &{0.281} &{\boldres{0.182}} &{\boldres{0.272}}  &{0.233} &{0.296} &{\boldres{0.213}} &{\boldres{0.272}} &{0.237} &{0.300} &{\boldres{0.231}} &{\boldres{0.295}}&{0.242} &{0.342} &{\boldres{0.226}} &{\boldres{0.319}} &{0.884} &{0.711} &{\boldres{0.603}} &{\boldres{0.545}} &{0.236} &{0.279} &{\boldres{0.214}} &{\boldres{0.245}} &{10.96}
    \\
    & {192}  &{0.233} &{0.261} & {\boldres{0.211}} & {\boldres{0.259}} &{0.261} &{0.301} &{\boldres{0.207}} &{\boldres{0.275}}  &{0.260} &{0.316} &{\boldres{0.234}} &{\boldres{0.292}} &{0.265} &{0.321} &{\boldres{0.261}} &{\boldres{0.318}}&{0.285} &{0.380} &{\boldres{0.245}} &{\boldres{0.366}} &{0.834} &{0.692} &{\boldres{0.682}} &{\boldres{0.563}} &{0.227} &{0.287} &{0.241} &{0.290} &{7.64}
    \\
    & {336}  &{0.248} &{0.273} & {\boldres{0.222}}  &{\boldres{0.261}} &{0.271} &{0.299} &{\boldres{0.212}} &{\boldres{0.272}}  &{0.276} &{0.323} &{\boldres{0.247}} &{\boldres{0.301}} &{0.283} &{0.330}&{\boldres{0.277}} &{\boldres{0.325}}&{0.282} &{0.376} &{\boldres{0.246}} &{\boldres{0.350}} &{0.941} &{0.723} &{\boldres{0.739}} &{\boldres{0.588}} &{0.262} &{0.310} &{\boldres{0.246}} &{\boldres{0.307}} &{9.54}
    \\
    & {720}  &{0.249} &{0.275} &{\boldres{0.203}} &{\boldres{0.263}} &{0.267} &{0.293} &{\boldres{0.201}} &{\boldres{0.262}}  &{0.273} &{0.316} &{\boldres{0.244}} &{\boldres{0.291}} &{0.286} &{0.326} &{\boldres{0.281}} &{\boldres{0.322}}&{0.357} &{0.427} &{\boldres{0.304}} &{\boldres{0.410}} &{0.882} &{0.717} &{\boldres{0.801}} &{\boldres{0.642}} &{0.329} &{0.355} &{\boldres{0.279}} &{\boldres{0.329}} &{10.05}
    \\
    \cmidrule(lr){2-31}
    & {Avg}  & {0.233}  & {0.262}  & {\boldres{0.203}}  & {\boldres{0.251}}  & {0.255}  & {0.294}  & {\boldres{0.201}}  & {\boldres{0.270}}  & {0.261}  & {0.313}  & {\boldres{0.235}}  & {\boldres{0.289}}  & {0.268}  & {0.319}  & {\boldres{0.263}}  & {\boldres{0.315}}  & {0.292}  & {0.381}  & {\boldres{0.255}}  & {\boldres{0.361}}  & {0.885}  & {0.711}  & {\boldres{0.706}}  & {\boldres{0.585}}  & {0.264}  & {0.308}  & {\boldres{0.245}}  & {\boldres{0.293}}  & {9.65}
    \\
    \midrule
    \multirow{5}{*}{\rotatebox{90}{{PEMS04}}} 
    &  {96}  &{0.159} &{0.272} &{\boldres{0.124}} &{\boldres{0.237}} &{0.166} &{0.285} &{\boldres{0.111}} &{\boldres{0.220}} & {0.273} & {0.372} &{\boldres{0.180}} &{\boldres{0.292}} &{0.274} &{0.372} &{\boldres{0.211}} &{\boldres{0.311}}&{0.220} &{0.338} &{\boldres{0.219}} &{\boldres{0.336}} &{0.553} &{0.583} &{\boldres{0.398}} &{\boldres{0.471}} &{\boldres{0.123}} &{\boldres{0.236}} &{0.125} &{\boldres{0.236}} &{16.60}
    \\
    & {192}  &{0.182} &{0.290} &{\boldres{0.162}} &{\boldres{0.266}} &{0.196} &{0.166} &{\boldres{0.127}} &{\boldres{0.235}} & {0.308} & {0.395} &{\boldres{0.224}} &{\boldres{0.320}} &{0.307} &{0.392} &{\boldres{0.248}} &{\boldres{0.342}} &{0.313} &{0.419} &{\boldres{0.310}} &{\boldres{0.418}} &{0.938} &{0.772} &{\boldres{0.502}} &{\boldres{0.546}}&{0.142} &{0.252}&{\boldres{0.136}} &{\boldres{0.246}} &{12.48}
    \\
    & {336}  &{0.186} &{0.291} &{\boldres{0.166}} &{\boldres{0.268}}  &{0.204} &{0.312} &{\boldres{0.134}} &{\boldres{0.242}} & {0.283}& {0.373} &{\boldres{0.224}} &{\boldres{0.322}} &{0.285} &{0.375} &{\boldres{0.247}} &{\boldres{0.346}} &{\boldres{0.231}} &{\boldres{0.339}} &{0.232} &{\boldres{0.339}} &{0.953} &{0.772}&{\boldres{0.672}} &{\boldres{0.645}}&{0.155} &{0.263}&{\boldres{0.148}} &{\boldres{0.256}} &{13.12}
    \\
    & {720}  &{0.242} &{0.332} &{\boldres{0.203}} &{\boldres{0.298}} &{0.238} &{0.345} &{\boldres{0.148}} &{\boldres{0.259}} & {0.328}& {0.407} &{\boldres{0.258}} &{\boldres{0.352}} &{0.329} &{0.406} &{\boldres{0.286}} &{\boldres{0.376}} &{0.643} &{0.592} &{\boldres{0.605}} &{\boldres{0.577}} &{1.120} &{0.823}&{\boldres{0.879}} &{\boldres{0.747}}&{0.192} &{0.297} &{\boldres{0.162}}&{\boldres{0.272}}  &{14.83}
    \\
    \cmidrule(lr){2-31}
    & {Avg}  &{0.192}  &{0.296}  &{\boldres{0.164}}  &{\boldres{0.267}}  &{0.201}  &{0.277}  &{\boldres{0.130}}  &{\boldres{0.239}}  &{0.298}  &{0.387}  &{\boldres{0.222}}  &{\boldres{0.322}}  &{0.299}  &{0.386}  &{\boldres{0.248}}  &{\boldres{0.344}}  &{0.352}  &{0.422}  &{\boldres{0.342}}  &{\boldres{0.418}}  &{0.891}  &{0.738}  &{\boldres{0.613}}  &{\boldres{0.602}}  &{0.153}  &{0.262}  &{\boldres{0.143}}  &{\boldres{0.253}}  &{14.86}
    \\
    \midrule
    \multirow{5}{*}{\rotatebox{90}{{PEMS08}}} 
    & {96}  &{0.169} &{0.276} &{\boldres{0.109}} &{\boldres{0.216}} &{0.252} &{0.355} &{\boldres{0.124}} &{\boldres{0.239}} & {0.284}& {0.375} &{\boldres{0.198}} &{\boldres{0.300}} &{0.285} &{0.380} &{\boldres{0.220}} &{\boldres{0.328}} &{\boldres{0.221}} &{\boldres{0.325}} &{0.234} &{0.329} &{0.613} &{0.596}&{\boldres{0.444}} &{\boldres{0.508}}&{0.171} &{0.283}&{\boldres{0.165}} &{\boldres{0.274}} &{19.24}
    \\
    & {192}  &{0.188} &{0.288} &{\boldres{0.137}} &{\boldres{0.239}} &{0.322} &{0.385} &{\boldres{0.151}} &{\boldres{0.256}} & {0.336}& {0.409} &{\boldres{0.222}} &{\boldres{0.310}} &{0.335} &{0.409} &{\boldres{0.280}} &{\boldres{0.371}} &{\boldres{0.332}} &{\boldres{0.412}} &{0.340} &{0.420} &{1.111} &{0.829}&{\boldres{0.616}} &{\boldres{0.596}}&{\boldres{0.202}} &{0.304}&{\boldres{0.202}} &{\boldres{0.298}} &{20.35}
    \\
    & {336}  &{0.196} &{0.289} &{\boldres{0.148}} &{\boldres{0.243}} &{0.326} &{0.374} &{\boldres{0.170}} &{\boldres{0.265}} & {0.327}& {0.394} &{\boldres{0.266}} &{\boldres{0.337}} &{0.328} &{0.397} &{\boldres{0.284}} &{\boldres{0.363}} &{\boldres{0.247}} &{0.322} &{\boldres{0.247}} &{\boldres{0.321}} &{1.136} &{0.836}&{\boldres{0.716}} &{\boldres{0.648}}&{0.224} &{0.309}&{\boldres{0.210}} &{\boldres{0.299}} &{17.27}
    \\
    & {720}  &{0.235} &{0.320} &{\boldres{0.186}} &{\boldres{0.277}} &{0.388} &{0.422} &{\boldres{0.225}} &{\boldres{0.308}} & {0.372}& {0.430} &{\boldres{0.303}} &{\boldres{0.372}} &{0.373} &{0.431} &{\boldres{0.327}} &{\boldres{0.396}} &{0.564} &{0.556} &{\boldres{0.546}} &{\boldres{0.546}} &{1.409} &{0.948}&{\boldres{0.901}} &{\boldres{0.733}}&{0.247} &{0.324}&{\boldres{0.238}} &{\boldres{0.316}} &{16.12}
    \\
    \cmidrule(lr){2-31}
    & {Avg}  &{0.197}  &{0.293}  &{\boldres{0.145}}  &{\boldres{0.244}}  &{0.322}  &{0.384}  &{\boldres{0.168}}  &{\boldres{0.267}}  &{0.330}  &{0.402}  &{\boldres{0.247}}  &{\boldres{0.330}}  &{0.330}  &{0.404}  &{\boldres{0.278}}  &{\boldres{0.365}}  &{0.341}  &{0.404}  &{\boldres{0.342}}  &{\boldres{0.404}}  &{1.067}  &{0.802}  &{\boldres{0.669}}  &{\boldres{0.621}}  &{0.211}  &{0.305}  &{\boldres{0.204}}  &{\boldres{0.297}}  &{18.30}
    \\
    \bottomrule
  \end{tabular}
    \end{small}
  \end{threeparttable}
}
\end{table*}

\section{Experiments}
% \textcolor{purple}{Fine-tuning experiments? channel permutation experiments, random shuffled data.}

% \vspace{-0.2cm}
\subsection{Experimental Setup}
To evaluate the effectiveness of C-LoRA, we conduct extensive experiments using popular LTSF benchmarks and various backbones.

\noindent\textbf{Datasets.} Following the same setting as in \cite{liu2023itransformer,wu2022timesnet}, we consider 7 real-world time series datasets from various areas, including ETTh1, ETTm1, Weather, Electricity, Solar-Energy, PEMS04, and PEMS08.

\noindent\textbf{Backbone architectures.} 
We select a variety of neural forecasting architectures, including both CI and CD models. All of them are widely evaluated in previous work and have shown competitive performance. They are: Informer \cite{zhou2021informer}, Autoformer \cite{wu2021autoformer}, FEDformer \cite{zhou2022fedformer}, FreTS \cite{yi2024frequency}, RMLP \cite{li2023revisiting}, TSMixer \cite{chen2023tsmixer}, and iTransformer \cite{liu2023itransformer}.

% \noindent\textbf{Implementation.}
% The proposed model is implemented with Pytorch 1.9.1 on an NVIDIA RTX 2080Ti GPU.
% The hidden dimension $D$ is set to $32$.
% The number of MLP layers $L$ is set to $3$.
% For PEMS04, PEMS07, PEMS08, and PEMS-BAY datasets, we set the length of historical data $P$ to 12. For the Electricity dataset, we set $P=168$.
% For all datasets, we set the length of future data $F$ to 12.
% The learning rate is set to 0.001.

\vspace{-0.2cm}
\subsection{Performance Improvement by C-LoRA}
The results of long-term series forecasting are shown in Table \ref{tab:main}. As indicated, C-LoRA can consistently improve the performance of different backbones, including both CI models and CM models. For datasets with more prominent \textit{channel heterogeneity}, such as Solar and PeMS \cite{nie2023imputeformer}, the performance gains can be greater than $10\%$. For data with relatively small channels like ETT, we also observe improvements to some extent. Notably, models like RMLP can also benefit from C-LoRA even without explicit channel mixing modules.

\vspace{-0.1cm}
\begin{figure}[!htbp]
  \centering
  \captionsetup{skip=1pt}
  \includegraphics[width=1\columnwidth]{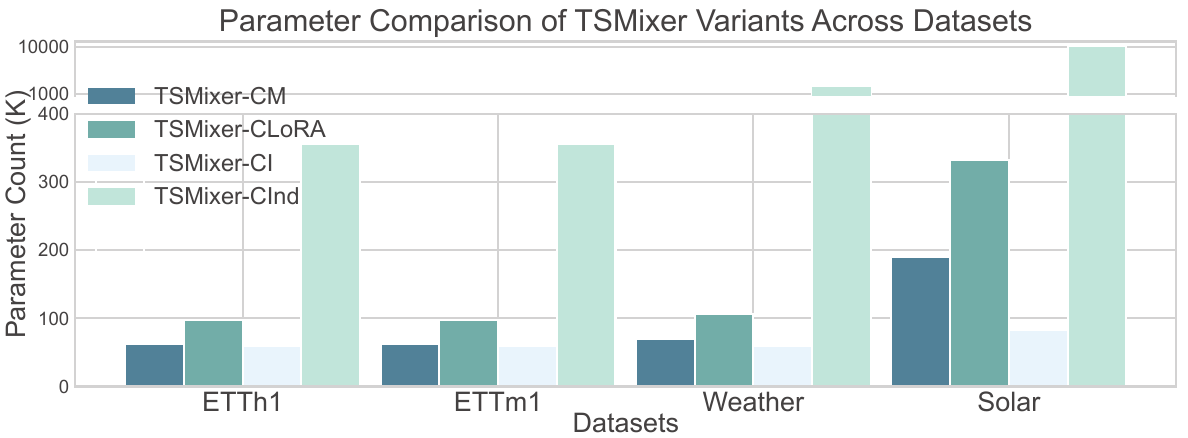}
  \caption{Parameter comparison of different strategies.}
  \label{fig:parameter_study}
\end{figure}
\vspace{-0.3cm}

\vspace{-0.3cm}
\subsection{Parameter Efficiency}
Due to the low-rank design, C-LoRA only introduces a few additional parameters to adapt the forecasting models. Fig. \ref{fig:parameter_study} compares it with both CI, CInd, and CD models. As can be seen, it significantly reduces the parameters of CInd models to consider individual treatment.
In addition, as C-LoRA contains only a few extra computations, it adds very limited additional computational overhead.

% \vspace{-0.3cm}
\begin{figure}[!htbp]
  \centering
  \captionsetup{skip=1pt}
  \includegraphics[width=1\columnwidth]{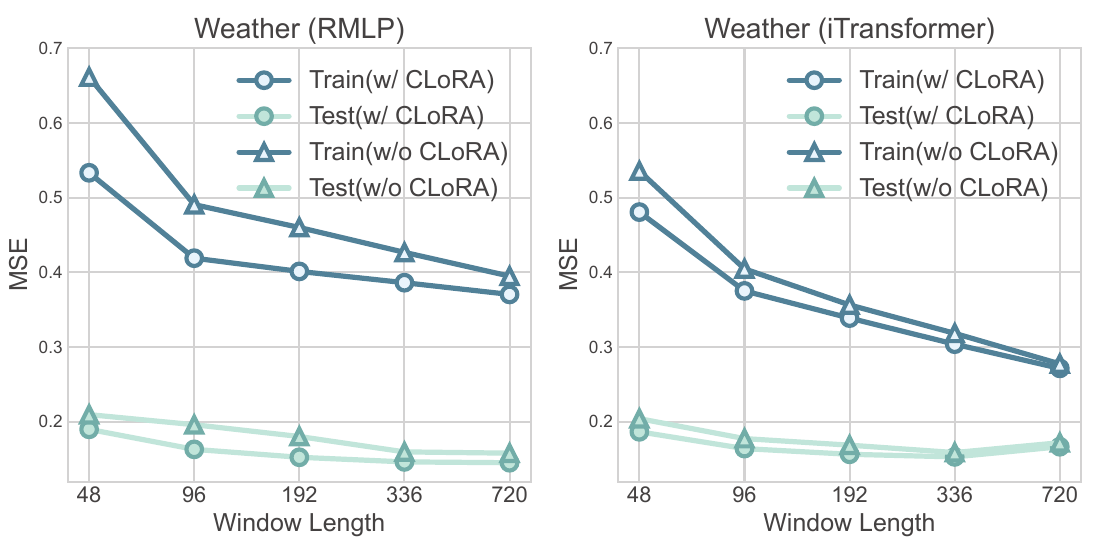}
  \caption{MSE under different lengths of look-back window.}
  \label{fig:window}
\end{figure}
\vspace{-0.2cm}

\subsection{Discussion and Analysis}
\noindent\textbf{Channel Mixing versus Channel Keeping.} Fig. \ref{fig:window} examines the training and test errors of both CM and CI models under different look-back windows. First, both of them benefit from a longer look-back window and the use of C-LoRA.
Second, the CD model has lower training errors, showing a larger capacity, however, can have larger generalization error than the CI model. Third, C-LoRA can narrow the performance gaps between training and test data.

\vspace{-0.3cm}
\begin{figure}[!htbp]
  \centering
  \captionsetup{skip=1pt}
  \includegraphics[width=1\columnwidth]{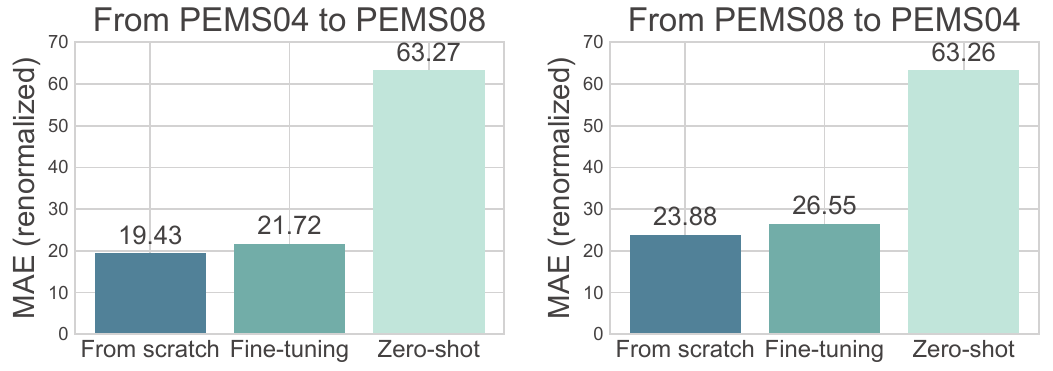}
  \caption{Transfer across datasets by fine-tuning C-LoRA.}
  \label{fig:finetune}
\end{figure}
\vspace{-0.3cm}

\noindent\textbf{Fine Tune C-LoRA.}
As shared patterns of time series can be captured by a CD model, a pretrained global model can be transferred to other datasets by efficient fine-tuning of C-LoRA. In Fig. \ref{fig:finetune}, we compare the results of \textbf{all} parameters trained from scratch, \textbf{only} fine-tuning C-LoRA on the target set, and the zero-shot model. Surprisingly, it can achieve desirable performance by only fine-tuning C-LoRA, indicating the flexibility of our hybrid channel strategy.

\vspace{-0.2cm}
\begin{figure}[!htbp]
  \centering
  \captionsetup{skip=1pt}
  \includegraphics[width=1\columnwidth]{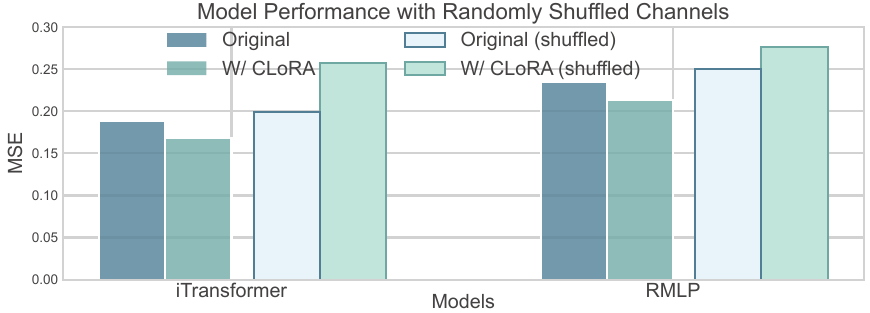}
  \caption{randomly shuffling the order of channels (Solar).}
  \label{fig:shuffle}
\end{figure}
\vspace{-0.2cm}

\noindent\textbf{Channel Identity Permutation.}
By randomly shuffling the order of channels, we can evaluate the importance of channel identity. It is observed that both CI and CD models show a performance drop after shuffling the order of channels.  In particular, models with C-LoRA have more pronounced error increases. This highlights that both CI and CD models can better preserve the channel identity information after being equipped with C-LoRA.

\begin{figure}[!htbp]
  \centering
  \captionsetup{skip=1pt}
  \includegraphics[width=1\columnwidth]{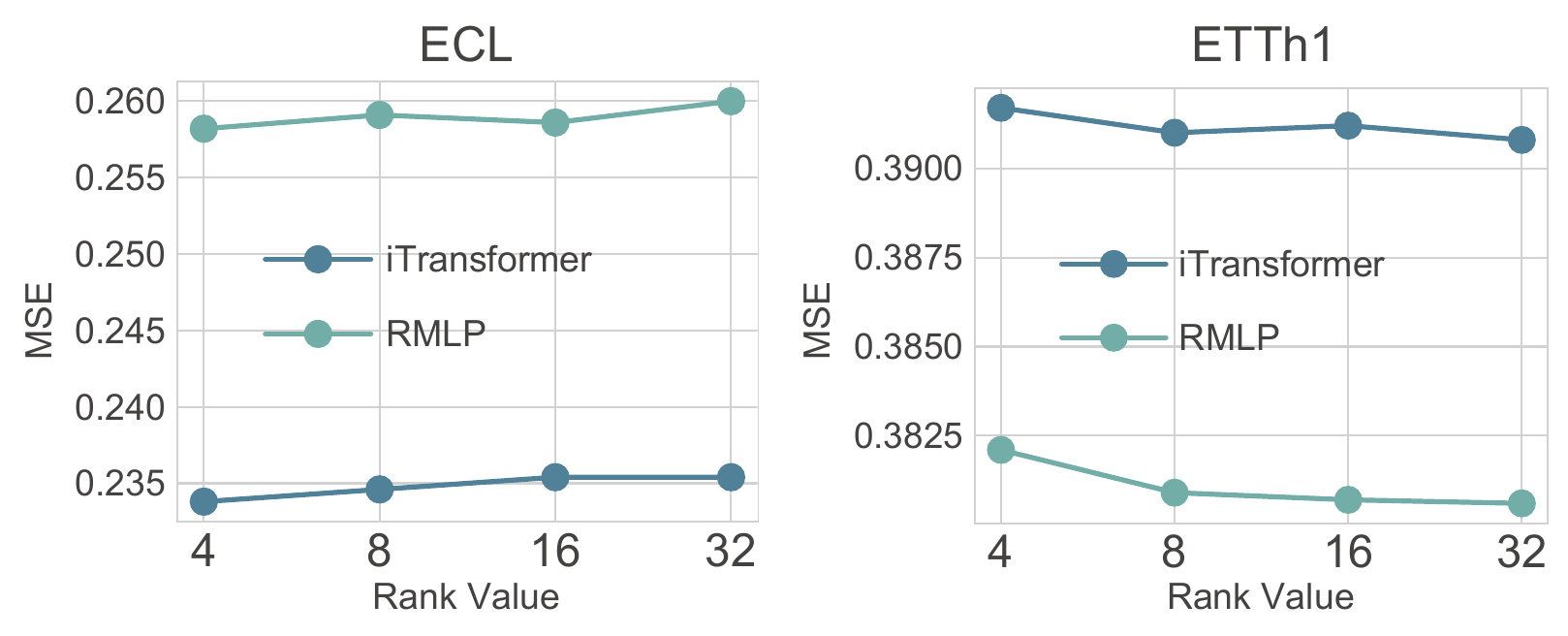}
  \caption{Model performance under different rank values.}
  \label{fig:rank_study}
\end{figure}

\noindent\textbf{Impact of Rank Selection.} Fig. \ref{fig:rank_study} studies the impact of different rank values. For time series having more channels with redundancy, a lower intrinsic rank is more beneficial \cite{nie2023correlating}. Conversely, a small data with heterogeneous channels prefer a large rank to ensure distinguishability. Both CI and CM methods show a similar trend.

\begin{figure}[!htbp]
  \centering
  \captionsetup{skip=1pt}
  \includegraphics[width=0.85\columnwidth]{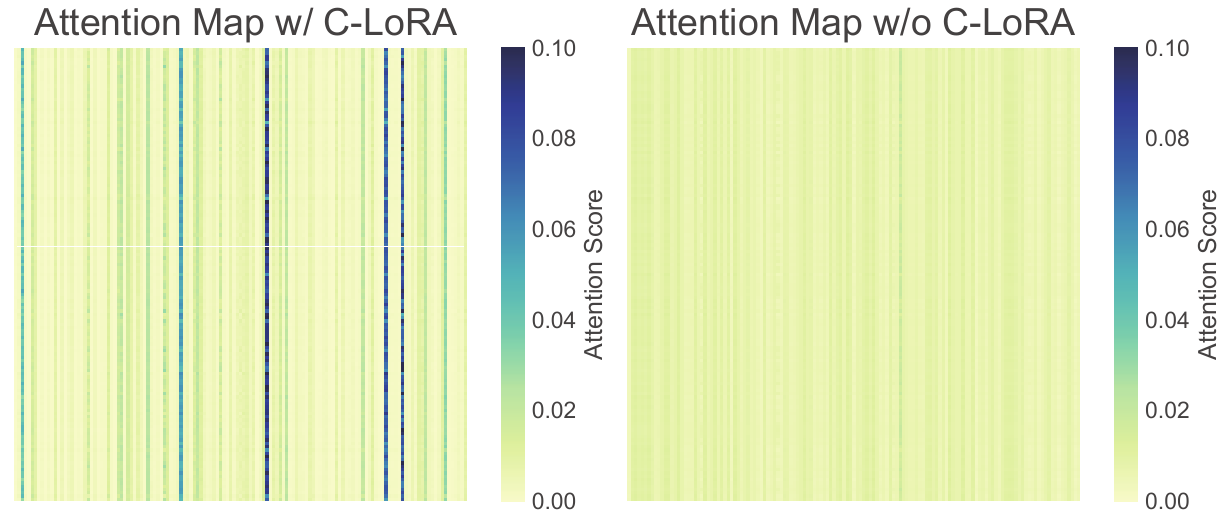}
  \caption{Attention scores w/ and w/o C-LoRA.}
  \label{fig:atten}
\end{figure}

\noindent\textbf{Case Study: Enhanced Attention Maps.} We visualize the attention maps of iTransformer of the same input with and without C-LoRA. Clearly, C-LoRA can help generate a attention map with more prominent patterns, focusing on a few critical channels. This indicates the role of preserving channel identity with C-LoRA.

\vspace{-0.2cm}
\section{Conclusion}
We introduce C-LoRA, a novel channel strategy to combine the strengths of both CI and CD strategies for long-term series forecasting. 
% By conditioning the channel-specific adapter on each series, it is aware of the channel identity to consider individual patterns. 
C-LoRA adapts to each channel with both channel-specific low-rank matrices and local patterns, bypassing the need for individual models. Potential cross-channel dependencies are preserved by a globally shared predictor. 
Extensive results indicate that it can consistently improve the performance of various backbones by a large margin. It is also efficient, flexible to fine-tune, and beneficial for the exploitation of channel identity.
Future work can evaluate its scalability in larger datasets, such as stock \cite{song2023stock} and traffic \cite{liu2024largest}.

\begin{acks}
The work was supported by research grants from the NSFC (52125208).
\end{acks}

\bibliographystyle{ACM-Reference-Format}
\normalem
\bibliography{references}

\end{document}